%% file: main.tex
\documentclass[runningheads]{llncs}
\usepackage[T1]{fontenc}
\usepackage{graphicx}
\usepackage{amsmath}
\usepackage{multirow}
\usepackage{amssymb}
\usepackage{cite}
\begin{document}
\title{MOSAIC: Orchestrating Collaborative Knowledge Tracing with Hierarchical Semantic Alignment}

\newif\ifarxiv
\arxivtrue  % arxiv version
% \arxivfalse  % official 

\author{Xinjin Li\inst{1}\textsuperscript{*, $\dagger$} \and
Mengyue Wang\inst{2}\textsuperscript{$\dagger$} \and
Yuzhen Lin\inst{3} \and
% Yue Wu\inst{4} \and
Pengbin Feng\inst{4} \and
Ziqi Sha\inst{5} \and
Yeyang Zhou\inst{6} \and
Yu Ma\inst{7}\textsuperscript{*}}

\authorrunning{X. Li et al.}
\titlerunning{MOSAIC}

\institute{Columbia University, New York, USA\\
\email{li.xinjin@columbia.edu} \and
University of California, Berkeley, Berkeley, USA\\
\email{cwang998@berkeley.edu} \and
School of Information Systems and Management, Carnegie Mellon University, Pittsburgh, PA, USA\\
\email{yuzhenl@alumni.cmu.edu} \and
% Tandon School of Engineering, New York University, New York, USA\\
% \email{y493423@gmail.com} \and
Department of Mathematics, University of Southern California, Los Angeles, USA\\
\email{pengbinf@alumni.usc.edu} \and
University of Massachusetts Amherst, Amherst, USA\\
\email{ziqisha@umass.edu} \and
Computer Science Department, UC San Diego, La Jolla, USA\\
\email{yeyang-zhou@ucsd.edu} \and
Carnegie Mellon Institute for Strategy \& Technology, Carnegie Mellon University, Pittsburgh, USA\\
\email{yuma13926@gmail.com}}
\maketitle              % typeset the header of the contribution

% --- Manually added footnotes for equal contribution and correspondence ---
\begingroup
  \renewcommand\thefootnote{} % Remove standard footnote numbering
  \footnotetext{\textsuperscript{*} Corresponding authors.}
  \footnotetext{\textsuperscript{$\dagger$} Equal contribution.}
\endgroup
% -------------------------------------------------------------------------

%
% \begin{abstract}
% The abstract should briefly summarize the contents of the paper in

% \keywords{First keyword  \and Second keyword \and Another keyword.}
% \end{abstract}
%
%
%

\input{sec/0_abstract}   
\input{sec/1_main}
{
    \small
    \bibliographystyle{splncs04}
    % \bibliography{main}

    \ifarxiv
    \bibliography{main,main_arxiv}
    \else
    \bibliography{main}
    \fi
}


\end{document}

%% file: sec/0_abstract.tex
\begin{abstract}
Knowledge Tracing (KT) is important for personalized education but traditionally suffers from two key limitations: a reliance on shallow ID-based representations that neglect semantic depth and a restriction to single-granularity mastery estimation that overlooks hierarchical knowledge dependencies. To address these challenges, we propose MOSAIC (Multi-granularity Online Semantic AI for Collaborative Knowledge), a novel framework that orchestrates LLM-driven semantic alignment with sequential modeling. Unlike methods that use LLMs solely as predictors, MOSAIC leverages a frozen LLM to generate dynamic, context-aware embeddings and hierarchical prediction prompts, explicitly capturing collaborative signals and peer interactions. Furthermore, we introduce a cross-granularity consistency objective that jointly regularizes mastery estimation across concept, topic-cluster, and global proficiency levels. Extensive experiments on ASSISTments, EdNet, and a newly collected large-scale MOOC dataset demonstrate that MOSAIC establishes new state-of-the-art results. Specifically, our method achieves AUC improvements of up to 3.4\% and Accuracy gains of up to 2.5 \% across all benchmarks. Notably, MOSAIC exhibits superior robustness in collaboration-rich environments and long-sequence scenarios (AUC 0.862 on MOOC), offering both high predictive precision and semantically grounded interpretability.
\keywords{Knowledge Tracing \and Large Language Models \and Multi-granularity Modeling \and Collaborative Learning \and Semantic Representation.}
\end{abstract}

%% file: sec/1_main.tex
\section{Introduction}

Knowledge tracing (KT) is a fundamental problem in AI for education, aiming to model how a learner's latent knowledge state evolves over time from sequential interaction data and to predict future performance for personalized intervention, tutoring, and curriculum adaptation \cite{corbett1994knowledge,piech2015dkt,shen2024survey}. Classical and neural KT methods have established strong performance on benchmark datasets and have become core components in intelligent tutoring systems. However, the learning environments faced by modern educational platforms are no longer limited to simple question--response logs. In realistic settings, student learning unfolds through heterogeneous and semantically rich signals, including attempts, hints, time-on-task, instructor feedback, and peer discussion. This shift raises a broader challenge for KT: beyond predicting the next response, a practical learner model should be able to track knowledge evolution under rich context and provide coherent estimates of mastery across different levels of abstraction.

Despite substantial progress, existing KT methods still face two persistent limitations. First, many models rely heavily on shallow ID-based representations for questions, concepts, and interactions. While effective for large-scale prediction, such representations often ignore the semantic content of exercises, the meaning of knowledge components, and the contextual information carried by collaborative learning signals. As a result, they remain limited in semantic expressiveness and are not well suited to realistic educational scenarios where textual and contextual evidence plays an important role. Second, most KT models operate primarily at a single granularity, typically the skill or concept level. This design overlooks the inherently hierarchical nature of learning, where fine-grained concept mastery is related to broader topic understanding and overall proficiency. Consequently, existing methods can achieve accurate local prediction while still lacking a coherent view of learner state across multiple knowledge levels \cite{abdelrahman2023knowledge,bai2024survey}.

Prior work only partially addresses these challenges. Probabilistic and psychometric formulations provide interpretable state-transition mechanisms, but they rely on simplified assumptions that limit their ability to capture complex learning behaviors \cite{corbett1994knowledge,lord2012applications}. Deep sequential KT models improve temporal modeling through recurrent, memory-based, and attention-based architectures, yet many still depend on ID-level embeddings and remain focused on a single granularity of mastery estimation \cite{piech2015dkt,zhang2017dkvmn,ghosh2020akt,lee2024monacobert}. Structure-aware and graph-based approaches further incorporate concept relations, but they often rely on predefined or weakly adaptive structures and do not naturally accommodate semantically rich collaborative context \cite{liu2021ekt,nakagawa2019gkt}. Meanwhile, large language models (LLMs) provide powerful semantic understanding and have shown strong potential in educational tasks such as feedback generation, tutoring dialogue, and explanation support \cite{kasneci2023chatgpt,achiam2023gpt}. Yet directly using LLMs as end-to-end predictors is not ideal for KT, where stable temporal state tracking and sequential inductive bias remain essential. These observations suggest that the key opportunity is not to replace KT with an LLM, but to integrate semantic understanding and temporal learner modeling in a principled way.

Motivated by this perspective, we propose \textbf{MOSAIC} (\textbf{M}ulti-granularity \textbf{O}nline \textbf{S}emantic \textbf{AI} for \textbf{C}ollaborative Knowledge), a unified framework for semantically grounded and hierarchically consistent knowledge tracing. The central idea is simple: we preserve a sequential KT backbone for modeling temporal knowledge evolution, while using a frozen LLM as a semantic alignment module to encode heterogeneous learning context. Specifically, MOSAIC leverages a frozen LLM to transform problem content, learning behaviors, and collaborative interaction text into context-aware semantic representations, which are then consumed by a sequential model for learner state tracking. On top of the shared latent state, MOSAIC jointly estimates mastery at three complementary levels---concept, topic-cluster, and global proficiency. To encourage coherent learner modeling across these levels, we further introduce a cross-granularity consistency objective that regularizes agreement between fine-grained and coarse-grained mastery estimates. In this way, MOSAIC combines the temporal robustness of conventional KT with the semantic richness of LLM-based representations, while moving beyond single-level prediction toward structured multi-level learner modeling.

We evaluate MOSAIC on ASSISTments, EdNet, and a large-scale university MOOC dataset that reflects collaboration-rich learning environments. Across benchmarks, MOSAIC consistently outperforms strong KT baselines, with particularly clear gains in settings involving long interaction sequences and rich collaborative context. These results suggest that combining semantic alignment with hierarchical inductive structure is a promising direction for next-generation KT systems. Our main contributions are summarized as follows:
\begin{itemize}
    \item We identify a key limitation of existing KT methods: although effective at temporal prediction, they remain insufficient for semantically grounded and hierarchically consistent learner modeling in realistic collaborative environments.
    \item We propose MOSAIC, a unified framework that couples frozen-LLM-based semantic alignment with sequential knowledge tracing to model learner mastery at the concept, topic-cluster, and global levels.
    \item We introduce a cross-granularity consistency objective that regularizes multi-level learner representations and show strong empirical gains on standard concept-level prediction across multiple benchmarks, especially in collaboration-rich and long-horizon learning scenarios.
\end{itemize}

\section{Related Work}

%arxiv
\ifarxiv

% \cite{li2023hong,chan2026adagar,meng2026dream,HUD,MEDIAN,Air-Know,zhang2025hypernetworks,liu2026agora,liu2025one,zhou2026comem,xiao2026metaucf,xiao2026protofedsp,ning2024physics,li2025identify,li2026sentinelvlametacognitivevlamodel,li2026vlaattcadaptivetesttimecompute,xu2024fakeshield,huang2025unishield,xu2026genshield,tao2025autopcr,zhang2025developing,guo2025quantized,qin2026explainable,qin2025interpretable,qin2025enhancing,yuan2025hekcl,chen2026diffgda,zhao2026generative,du2026point,zhang2026adaptive,jiao2026large,ji2025finestate,yang2026frequency,ji2026servimageimagegenerationediting,zang2024explanation,zang2025alleviating,zang2025compression,he2024priorvfi,he2026cinemattebackgroundmattingvirtual,dai2023neighbors,cheng2026resolvingrobustnessprecisiontradeofffinancial,11484350,cheng2026enhancingfinancialreportquestionanswering,zhang2026finsentllm,deng2026adaptive,shi2026multiscenario,jiang2026drpdistilledreasoningpruning,jiang2026scribe,zhang2026optimalteacherpersonalizeddata,chen2026rag,lin2026reflect,tian2024mmrec,wang2025dlrrec,10.1145/3764926.3771951,10.1145/3814573.3814949,feng2026leveraging,feng2026prism,feng2023unsupervised,wang2026safeskillscollidemeasuring,qian2026relevantwarrantedevidenceforcecalibration,han2026interpretable,zhang2026bankruptcy}
\subsection{Broader Impacts of LLMs and Representation Learning}
Recent advances in Large Language Models (LLMs) and representation learning have profoundly influenced a wide spectrum of domains, establishing a foundational paradigm for multi-modal and adaptive intelligence. In embodied intelligence and autonomous systems, researchers have significantly enhanced dynamic environmental interaction and multimodal understanding through scene perception, continuous adaptation, metacognitive coordination, and safe skill execution~\cite{li2023hong,chan2026adagar,meng2026dream,HUD,MEDIAN,Air-Know,zhou2026comem,li2025identify,li2026sentinelvlametacognitivevlamodel,li2026vlaattcadaptivetesttimecompute,du2026point,zhang2026adaptive,jiao2026large,ji2025finestate,yang2026frequency,ji2026servimageimagegenerationediting,he2026cinemattebackgroundmattingvirtual,deng2026adaptive,shi2026multiscenario,10.1145/3764926.3771951,10.1145/3814573.3814949,wang2026safeskillscollidemeasuring}, while concurrent work in federated, continual, and personalized representation learning investigates heterogeneous adaptation, task-conditioned parameter generation, and graph-based evolution for user-specific intelligence~\cite{zhang2025hypernetworks,liu2025one,xiao2026metaucf,xiao2026protofedsp,ning2024physics,yuan2025hekcl,chen2026diffgda,feng2026prism,feng2023unsupervised,feng2026leveraging}. Concurrently, existing efforts have accelerated developments in trustworthy AI, model robustness, and explainability by mitigating hallucinations and establishing rigorous defense, counterfactual reasoning, reflection, and interpretation frameworks~\cite{xu2024fakeshield,huang2025unishield,xu2026genshield,qin2026explainable,qin2025interpretable,qin2025enhancing,zang2024explanation,zang2025alleviating,zang2025compression,lin2026reflect,han2026interpretable}, while expanding into life sciences for multimodal genomic analysis and phenotype recognition~\cite{tao2025autopcr,zhang2025developing,guo2025quantized}. Furthermore, efficient reasoning pipelines and tool-augmented ecosystems have emerged to optimize scalability through reasoning pruning, structured supervision, and reliable evidence calibration~\cite{liu2026agora,jiang2026drpdistilledreasoningpruning,jiang2026scribe,zhang2026optimalteacherpersonalizeddata,chen2026rag,qian2026relevantwarrantedevidenceforcecalibration}, complemented by generative paradigms and multimodal fusion that advance graph modeling, personalization, and rich semantic understanding in recommender systems~\cite{zhao2026generative,tian2024mmrec,wang2025dlrrec} and computer vision tasks~\cite{he2024priorvfi,feng2026leveraging,ji2026servimageimagegenerationediting}. Finally, these foundation models have demonstrated substantial socioeconomic impact within finance and decision intelligence, driving innovations in financial question answering, sentiment forecasting, bankruptcy prediction, and interpretable market analysis beyond traditional benchmarks~\cite{dai2023neighbors,cheng2026resolvingrobustnessprecisiontradeofffinancial,11484350,cheng2026enhancingfinancialreportquestionanswering,zhang2026finsentllm,han2026interpretable,zhang2026bankruptcy}.
\fi

\subsection{The Evolution of Sequential Knowledge Tracing}
Knowledge tracing has evolved from classical probabilistic student models to highly expressive deep sequential architectures. Early methods, such as Bayesian Knowledge Tracing~\cite{corbett1994knowledge}, provide interpretable state transitions but rely on overly simplified assumptions regarding learning dynamics, although related Bayesian adaptive-control ideas have also been explored in sequential decision-making settings \cite{11059994}. The introduction of Deep Knowledge Tracing~\cite{piech2015dkt} demonstrated that recurrent networks could effectively map raw interaction sequences to future performance. Subsequent architectures have heavily refined this temporal modeling through memory and attention mechanisms, prominently including Dynamic Key-Value Memory Networks~\cite{zhang2017dkvmn}, AKT~\cite{ghosh2020akt}, and MonaCoBERT~\cite{lee2024monacobert}. More recent iterations further constrain these sequences by incorporating question difficulty~\cite{liu2024qdkt} or hierarchical session structures~\cite{ke2024hitskt}, with related work also exploring adaptive context-length optimization in multi-agent reinforcement learning settings \cite{duanadaptive}. However, despite achieving strong predictive accuracy, these sequential baselines remain bottlenecked by their reliance on shallow, ID-based representations and short behavioral features. Because they fundamentally optimize for single-granularity, next-response prediction, they lack the semantic depth required to model collaboration-rich environments or complex peer interactions.

\subsection{Structure- and Hierarchy-Aware Knowledge Tracing}
To move beyond the limitations of pure ID-based modeling, a parallel line of research injects content and structural priors into the KT pipeline. Models like EERNN~\cite{su2018eernn} and EKT~\cite{liu2021ekt} exploit exercise text to improve question representations, while Graph-based Knowledge Tracing~\cite{nakagawa2019gkt} formalizes concept dependencies using graph neural networks. More recently, Relation-Aware KT~\cite{pandey2020rkt} combined text-derived exercise relations with forgetting behaviors to improve predictive robustness. While these studies validate the importance of relational inductive biases, they rely on predefined, static schemas and restrict semantic enrichment to the individual question or concept level. Consequently, they are inherently ill-equipped to process unstructured, dynamic collaborative signals---such as open-ended peer discussions---and rarely produce jointly constrained, multi-level mastery estimates across fine-grained concepts, topic clusters, and overall proficiency.

\subsection{Large Language Models as Semantic Engines in KT}
Recognizing the limitations of static representations, recent literature has begun integrating Large Language Models (LLMs) into educational modeling. Current approaches generally fall into three paradigms: text-aware representation learning (e.g., LKT~\cite{lee2024lkt}), LLM-driven structural generation (e.g., SINKT~\cite{fu2024sinkt}, which builds heterogeneous concept-question graphs), and LLM-centered prediction or profile-based forecasting (e.g., LLM-KT~\cite{wang2025llmkt} and CIKT~\cite{li2025cikt}, which respectively use plug-and-play instructions and analyst-predictor loops). While these works successfully demonstrate the value of language-model priors, they primarily focus on text-aware representation learning, inductive or cold-start generalization, or LLM-centered prediction and profile generation. They do not address the specific challenge resolved in this work: orchestrating collaborative, multi-granularity knowledge tracing. This gap necessitates MOSAIC, which uniquely isolates the LLM as a frozen semantic enhancer and prompt constructor. This design injects deep semantic alignment and cross-granularity consistency into the pipeline while preserving the efficiency and temporal stability of a dedicated sequential backbone.

\section{Method}

\begin{figure*}[t]
  \centering
  \includegraphics[width=\linewidth]{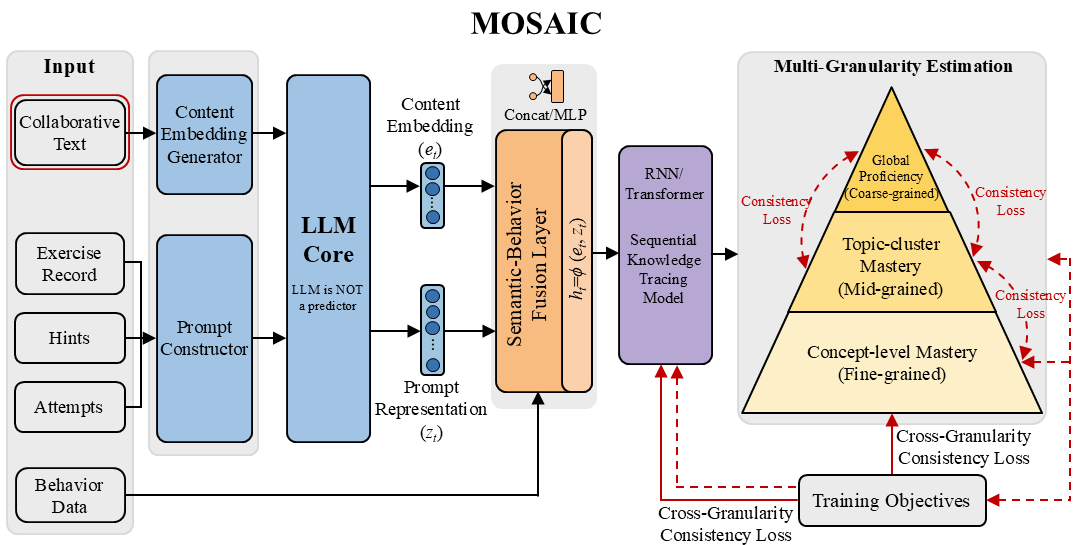}
  \caption{The overall architecture of the MOSAIC framework. The model utilizes a frozen LLM to function as a semantic enhancer and prompt constructor, processing heterogeneous inputs including collaborative text and exercise records. The right panel illustrates the hierarchical multi-granularity estimation, where concept-level, topic-cluster, and global proficiency states are jointly optimized via cross-granularity consistency losses.}
  \label{fig:mosaic_arch}
\end{figure*}

\subsection{Problem Formulation and Overview}

We consider the knowledge tracing problem in a collaborative learning environment. For each student $u$, we observe a chronologically ordered interaction sequence
$
\mathcal{S}^{(u)} = \{ (q_t, r_t, b_t, c_t) \}_{t=1}^{T_u},
$
where $q_t$ denotes the exercised question at time step $t$, $r_t \in \{0,1\}$ is the student's response correctness, $b_t$ represents auxiliary learning behaviors (\textit{e.g.}, attempt count or time-on-task), and $c_t$ denotes the associated knowledge concept or skill tag. In addition, each interaction may be accompanied by collaborative textual context $x_t$, such as peer discussion or instructor feedback, reflecting social learning signals.

The goal of knowledge tracing is to estimate the student's latent knowledge state over time and predict future performance. Unlike conventional KT, which focuses on a single granularity, MOSAIC jointly models learner mastery at multiple knowledge levels. Specifically, at each time step $t$, the model outputs mastery probabilities at the concept level $\hat{y}_t^{(c)}$, topic-cluster level $\hat{y}_t^{(g)}$, and global proficiency level $\hat{y}_t^{(u)}$:
$
(\hat{y}_t^{(c)}, \hat{y}_t^{(g)}, \hat{y}_t^{(u)}) = f_\theta(\mathcal{S}^{(u)}_{\le t}, x_{\le t}),
$
where $f_\theta$ denotes the proposed MOSAIC model parameterized by $\theta$. The topic-cluster level corresponds to a higher-level grouping of related concepts, while the global level reflects an overall estimate of the student's learning proficiency.

As illustrated in Figure~\ref{fig:mosaic_arch}, MOSAIC follows a simple design principle. A frozen LLM is used to provide semantically grounded, context-aware representations from heterogeneous learning signals, while a lightweight sequential KT backbone models temporal knowledge evolution. On top of the shared latent state, three prediction heads estimate mastery at different granularities, and a consistency objective encourages coherent knowledge estimation across fine-grained and coarse-grained levels.

\subsection{Hierarchical Semantic Alignment with Frozen LLM}

A key limitation of existing KT models is their reliance on discrete ID-based embeddings for questions and concepts, which lack semantic expressiveness. To address this issue, MOSAIC leverages a large language model (LLM) as a semantic enhancement module that transforms heterogeneous learning signals into context-aware representations.

For each time step $t$, we construct a textual description $\tau_t$ that summarizes the current learning context, including the exercised question $q_t$, its associated concept $c_t$, the student's response $r_t$, auxiliary behaviors $b_t$, and available collaborative interaction text $x_t$. The LLM encoder then maps this description to a dense semantic representation:
$
\mathbf{e}_t = \mathrm{LLM}_{\text{enc}}(\tau_t) \in \mathbb{R}^d,
$
where $\mathbf{e}_t$ serves as a dynamic knowledge embedding that captures semantic relations among questions, concepts, and interactions.

In addition to dynamic embeddings, MOSAIC uses the LLM to construct prediction prompts that explicitly encode hierarchical knowledge context and recent learning history. Formally, we define a prompt template $\pi(\cdot)$ that conditions on the interaction prefix up to time $t$, producing a structured prompt $p_t = \pi(\mathcal{S}_{\le t}, x_{\le t})$. The LLM processes this prompt to generate a prompt-aware representation:
$
\mathbf{z}_t = \mathrm{LLM}_{\text{prompt}}(p_t),
$
which provides high-level semantic guidance for downstream knowledge state modeling. Importantly, the LLM is not used to directly predict correctness; instead, the generated embeddings $\mathbf{e}_t$ and prompt representations $\mathbf{z}_t$ are treated as auxiliary semantic inputs. This design decouples semantic understanding from sequential prediction, allowing MOSAIC to retain the robustness of conventional KT models while benefiting from the rich representational capacity of LLMs.

\subsection{Sequential Knowledge State Modeling}

Given the LLM-generated semantic embedding $\mathbf{e}_t$ and prompt-aware representation $\mathbf{z}_t$ at each time step, MOSAIC models the temporal evolution of a student's knowledge state using a sequential encoder. We first combine semantic and behavioral information through a fusion function $\phi(\cdot)$:
$
\mathbf{h}_t = \phi(\mathbf{e}_t, \mathbf{z}_t),
$
where $\mathbf{h}_t$ represents the enriched interaction representation at time $t$. In practice, $\phi(\cdot)$ can be instantiated as concatenation followed by a linear projection.

\begin{figure*}[t]
  \centering
  \includegraphics[width=\linewidth]{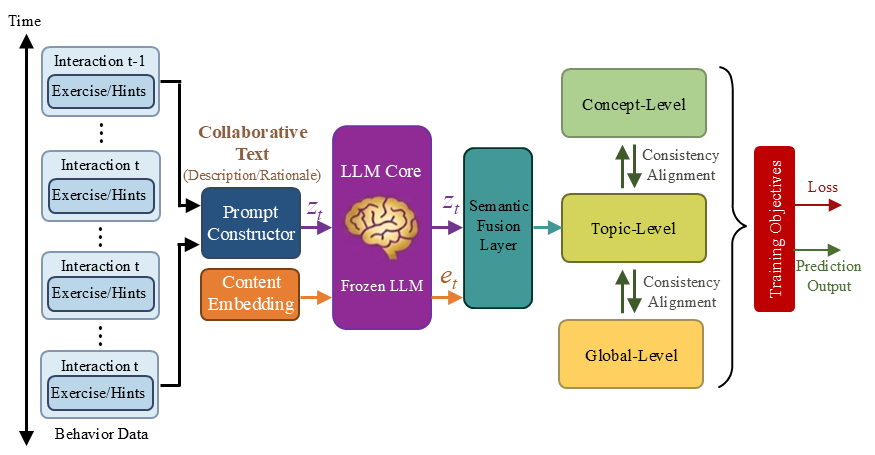}
  \caption{The sequential processing workflow and multi-granularity semantic alignment in MOSAIC. The diagram illustrates how interaction history and collaborative text are transformed by the frozen LLM into dynamic latent states ($\mathbf{z}_t, \mathbf{e}_t$) over time. The right section highlights the \textbf{consistency alignment mechanism}, which enforces logical coherence between fine-grained concept mastery, mid-grained topic mastery, and coarse-grained global proficiency during the sequential modeling process.}
  \label{fig:mosaic_time}
\end{figure*}

The sequence $\{\mathbf{h}_t\}_{t=1}^{T_u}$ is then fed into a sequential model to capture temporal dependencies in the student's learning trajectory:
$
\mathbf{s}_t = \mathrm{SeqModel}(\mathbf{h}_1, \ldots, \mathbf{h}_t),
$
where $\mathbf{s}_t$ denotes the latent knowledge state at time $t$. The sequential model can be implemented using a recurrent neural network or a transformer-based architecture, enabling MOSAIC to model both short-term learning dynamics and long-range knowledge accumulation.

Compared with conventional KT models that rely solely on ID-based inputs, the latent state $\mathbf{s}_t$ in MOSAIC is conditioned on semantically rich and context-aware representations, allowing the model to better capture nuanced learning patterns and the influence of collaborative interactions over time. This stage preserves the temporal inductive bias of knowledge tracing while enriching the state dynamics with semantic signals from the frozen LLM.

\subsection{Multi-Granularity Mastery Estimation}

Based on the latent knowledge state $\mathbf{s}_t$, MOSAIC jointly estimates learner mastery at multiple knowledge granularities. Specifically, we employ three prediction heads to model concept-level, topic-cluster-level, and global proficiency mastery. For each time step $t$, the corresponding mastery probabilities are computed as
$
\hat{y}_t^{(c)} = \sigma(\mathbf{W}_c \mathbf{s}_t), \quad
\hat{y}_t^{(g)} = \sigma(\mathbf{W}_g \mathbf{s}_t), \quad
\hat{y}_t^{(u)} = \sigma(\mathbf{W}_u \mathbf{s}_t),
$
where $\sigma(\cdot)$ denotes the sigmoid function, and $\mathbf{W}_c$, $\mathbf{W}_g$, and $\mathbf{W}_u$ are learnable parameters for the concept-, topic-, and global-level predictors, respectively.

This design allows MOSAIC to produce fine-grained and coarse-grained mastery estimates from a shared latent state. Rather than treating higher-level signals as auxiliary outputs, MOSAIC explicitly models hierarchical knowledge states within a unified prediction framework.

\subsection{Cross-Granularity Consistency Optimization}

The consistency alignment mechanism is illustrated in Figure~\ref{fig:mosaic_time}. The model is trained with a standard prediction loss $\mathcal{L}_{\text{pred}}$, defined as the binary cross-entropy between the predicted mastery and observed student responses at the concept level. To ensure coherent knowledge estimation across granularities, we further introduce a cross-granularity consistency loss that regularizes the agreement between fine-grained and coarse-grained predictions:
$
\mathcal{L}_{\text{cons}} =
\sum_t \left(
\lVert \hat{y}_t^{(c)} - \hat{y}_t^{(g)} \rVert_2^2
+
\lVert \hat{y}_t^{(g)} - \hat{y}_t^{(u)} \rVert_2^2
\right).
$
This objective encourages concept-level mastery to align with its corresponding topic-cluster estimate, and topic-level mastery to remain consistent with global proficiency, while still allowing flexibility for local variations.

The final training objective is given by
$
\mathcal{L} = \mathcal{L}_{\text{pred}} + \lambda \, \mathcal{L}_{\text{cons}},
$
where $\lambda$ controls the strength of cross-granularity regularization. This joint objective enables MOSAIC to produce stable and interpretable mastery estimates across multiple knowledge levels.

\section{Experiments}

\subsection{Experimental Setup}
\label{sec:exp_setup}

\noindent{\textbf{Datasets}} We evaluate MOSAIC on three educational datasets. ASSISTments contains student--problem interaction logs with binary correctness labels and expert-annotated skill tags, and is a standard benchmark for concept-level knowledge tracing. EdNet provides large-scale student interaction sequences with rich temporal information and fine-grained learning behaviors, making it suitable for evaluating knowledge tracing under long and dense learning trajectories. In addition, we use a Chinese University MOOC dataset collected from public online courses, which includes problem-solving records, detailed learning behaviors, expert-provided knowledge tags, and collaborative interaction text such as discussion posts and peer comments. This dataset is particularly useful for evaluating knowledge tracing in collaboration-rich learning environments.

\noindent{\textbf{Task Definition and Evaluation Protocol}} Our primary task is standard concept-level next-response prediction: given a student's interaction history up to time step $t$, the model predicts the correctness of the next interaction. For each dataset, student interaction sequences are ordered chronologically and split into training, validation, and test sets at the student level with a ratio of 8:1:1 to avoid information leakage. Model selection and hyperparameter tuning are performed on the validation set.

\noindent{\textbf{Baselines}} We compare MOSAIC against representative knowledge tracing baselines spanning classical probabilistic modeling, neural sequential modeling, and recent attention-based architectures, including BKT, DKT, DKVMN, AKT, and MonaCoBERT. These baselines provide strong reference for evaluating whether the proposed semantic alignment and hierarchical design yield consistent improvements over existing KT approaches.

\noindent{\textbf{Evaluation Metrics}} We report \textit{Area Under the ROC Curve (AUC)} and \textit{Accuracy} as the evaluation metrics. AUC serves as the primary metric because it measures ranking quality independent of a fixed decision threshold, while Accuracy is reported as a complementary thresholded metric.

\noindent{\textbf{Implementation Details}} The sequential knowledge tracing component is implemented using a transformer-based encoder with 2 layers, 4 attention heads, and a hidden dimension of $d=256$. For semantic enhancement, we employ Qwen2.5-7B-Instruct as a frozen LLM to generate dynamic knowledge embeddings and prediction prompts for both English (ASSISTments, EdNet) and Chinese (MOOC) inputs. To ensure efficiency, all LLM outputs are extracted offline, cached, and reused during training and inference, with no gradient back-propagation through the LLM.

\subsection{Main Results on Standard Knowledge Tracing}
\label{sec:overall_results}

\begin{table}[t]
\centering
\caption{Concept-level next-response prediction performance on ASSISTments and EdNet.}
\label{tab:overall_standard}
\begin{tabular}{lcc|cc}
\hline
\multirow{2}{*}{Model} & \multicolumn{2}{c|}{ASSISTments} & \multicolumn{2}{c}{EdNet} \\
 & AUC & Acc & AUC & Acc \\
\hline
BKT & 0.731 & 0.681 & 0.714 & 0.667 \\
DKT & 0.789 & 0.742 & 0.803 & 0.758 \\
DKVMN & 0.812 & 0.761 & 0.826 & 0.774 \\
AKT & 0.835 & 0.783 & 0.842 & 0.791 \\
MonaCoBERT & 0.847 & 0.794 & 0.856 & 0.802 \\
\hline
MOSAIC (ours) & \textbf{0.881} & \textbf{0.818} & \textbf{0.886} & \textbf{0.821} \\
\hline
\end{tabular}
\end{table}

\begin{table}[t]
\centering
\caption{Concept-level next-response prediction performance on the Chinese University MOOC dataset.}
\label{tab:overall_mooc}
\begin{tabular}{lcc}
\hline
Model & AUC & Acc \\
\hline
DKT & 0.771 & 0.723 \\
DKVMN & 0.796 & 0.744 \\
AKT & 0.821 & 0.768 \\
MonaCoBERT & 0.834 & 0.776 \\
\hline
MOSAIC (ours) & \textbf{0.862} & \textbf{0.801} \\
\hline
\end{tabular}
\end{table}

The gains over the strongest baselines are substantial rather than marginal, with AUC improvements of 3.4\% on ASSISTments and 3.0\% on EdNet. This pattern suggests that the advantage does not come from temporal modeling alone, since AKT and MonaCoBERT already provide competitive sequence modeling capacity. A more plausible explanation is that MOSAIC augments the sequential backbone with semantically grounded representations and prompt-aware hierarchical guidance, allowing it to move beyond shallow ID-based interaction encoding. By contrast, existing KT baselines mainly optimize single-granularity next-response prediction and therefore have limited ability to exploit rich problem semantics and contextual learning signals.

The larger margin on the MOOC benchmark is consistent with this interpretation. Because MOOC contains richer behavioral context and collaborative interaction text, it better exposes the weakness of purely ID-driven KT models and highlights the benefit of semantically aligned modeling. This result suggests that MOSAIC is not only a stronger generic predictor, but is particularly effective when the learning environment contains heterogeneous and collaboration-rich signals.

\subsection{Ablation Study}
\label{sec:ablation}

\begin{table}[t]
\centering
\caption{Ablation results on ASSISTments.}
\label{tab:ablation}
\begin{tabular}{lcc}
\hline
Variant & AUC & Acc \\
\hline
MOSAIC (full) & \textbf{0.881} & \textbf{0.818} \\
\hline
w/o LLM-driven embeddings & 0.852 & 0.800 \\
w/o prompt-based prediction & 0.863 & 0.802 \\
w/o collaborative text & 0.860 & 0.799 \\
w/o consistency loss & 0.865 & 0.811 \\
\hline
\end{tabular}
\end{table}

We conduct ablation studies on ASSISTments to quantify the contribution of each major component in MOSAIC. All ablated variants share the same sequential backbone and training protocol, differing only in the removed component.

Removing the LLM-driven semantic embeddings causes the largest drop, reducing AUC from 0.881 to 0.852. This suggests that the key limitation of prior KT models is not simply insufficient sequence modeling, but the lack of semantically expressive representations. Without this component, the model falls back toward the shallow interaction encoding regime of conventional ID-based KT methods. Removing prompt-based prediction also degrades performance, indicating that semantic embeddings alone are insufficient and that prompt-aware hierarchical guidance provides additional structure for organizing learning history and supporting prediction.

Excluding collaborative text further hurts performance, showing that the gain is not merely from adding an LLM-derived feature pipeline, but specifically from MOSAIC's ability to extract useful information from social learning signals. Removing the cross-granularity consistency loss also reduces both AUC and Accuracy, suggesting that consistency regularization improves optimization stability and helps preserve coherent hierarchical structure in the learned representations. Taken together, the ablations explain why competing methods underperform: they may model temporal dependencies well, but they lack one or more components needed to align semantic context, collaborative evidence, and hierarchical learner-state estimation within a unified framework.

\subsection{Analysis in Challenging Settings}
\label{sec:collab_long}

\begin{table}[t]
\centering
\caption{Effect of collaborative interaction modeling on the MOOC dataset.}
\label{tab:collaboration}
\begin{tabular}{lcc}
\hline
Model & AUC & $\Delta$AUC \\
\hline
AKT (w/o collaboration) & 0.821 & -- \\
AKT (with collaboration) & 0.834 & +1.3 \\
\hline
MOSAIC (w/o collaboration) & 0.845 & -- \\
MOSAIC (full) & \textbf{0.862} & \textbf{+1.7} \\
\hline
\end{tabular}
\end{table}

\begin{table}[t]
\centering
\caption{Performance (AUC) under different sequence lengths on EdNet.}
\label{tab:long_sequence}
\begin{tabular}{lccc}
\hline
Sequence Length & AKT & MonaCoBERT & MOSAIC \\
\hline
$\leq$100 & 0.850 & 0.864 & \textbf{0.876} \\
100--300 & 0.840 & 0.848 & \textbf{0.889} \\
$\geq$300 & 0.810 & 0.827 & \textbf{0.894} \\
\hline
\end{tabular}
\end{table}

We further analyze MOSAIC in two challenging yet realistic settings: collaboration-rich learning environments and long interaction sequences.

On the MOOC dataset, collaborative interaction modeling benefits both AKT and MOSAIC, but the gain is larger for MOSAIC. This suggests that collaborative text is not automatically useful; it becomes more helpful only when the model can semantically align unstructured peer interactions with the student's evolving knowledge state. In this sense, MOSAIC converts collaboration from noisy side information into task-relevant semantic evidence more effectively than standard KT models.

A similar pattern appears in the long-sequence analysis on EdNet. As sequence length increases, baseline performance declines, whereas MOSAIC remains strong and achieves its largest advantage in the longest-sequence regime. This suggests that semantically enriched representations and prompt-based guidance preserve informative context over extended learning histories, reducing the brittleness of purely ID-driven sequential modeling when temporal dependencies become long and complex. The fact that MOSAIC's relative advantage grows with sequence length further supports the claim that its gains come from better context modeling rather than a narrow improvement in short-horizon prediction.

Overall, these results indicate that MOSAIC is particularly effective in realistic KT scenarios characterized by richer social context and longer temporal dependencies. More importantly, the advantage appears precisely in the settings where the proposed semantic alignment and hierarchical inductive structure should matter most, strengthening the empirical support for the central design claim of the model.

\subsection{Qualitative Discussion on Interpretability}
\label{sec:interpretability}

Beyond predictive performance, MOSAIC is designed to provide a more structured view of learner states through its multi-granularity architecture and semantically informed representations. Unlike conventional KT models that output only a single mastery score, MOSAIC produces concept-level, topic-cluster-level, and global proficiency predictions from a shared latent state, which may facilitate a more organized analysis of learning progress.

The cross-granularity consistency objective is intended to encourage coherence across these levels of prediction, while the semantically enriched representations provide a principled way to connect model outputs with problem content, behavioral context, and collaborative interactions. Compared with conventional single-granularity KT models, this design offers a potentially more structured interface for analyzing learner dynamics and supporting downstream feedback.

Because MOSAIC conditions its representations on problem content, response outcomes, and collaborative interaction text, changes in the predicted learner state can in principle be associated with meaningful learning events, such as persistent difficulties on a specific concept or improvements following peer discussion. In this sense, MOSAIC may provide a semantically grounded interface for interpreting learner dynamics in practical educational applications that require both prediction quality and structured feedback.

\section{Conclusion}

We addressed a central limitation of knowledge tracing: existing methods often struggle to jointly capture semantic richness, hierarchical knowledge structure, and collaborative learning signals. To this end, we presented \textbf{MOSAIC}, a unified framework that combines frozen-LLM-based semantic alignment with a sequential knowledge tracing backbone, enabling student modeling that is both semantically informed and temporally grounded.

Rather than treating the LLM as an end-to-end predictor, MOSAIC uses dynamic semantic embeddings and prompt-aware representations to enhance sequential modeling, while jointly estimating concept-level, topic-cluster-level, and global proficiency states under a cross-granularity consistency objective. This design preserves the temporal inductive bias of knowledge tracing while extending it beyond shallow ID-based representations. Empirically, MOSAIC achieves consistent improvements over strong baselines on ASSISTments, EdNet, and a large-scale Chinese University MOOC dataset, with gains of up to 3.4 \% AUC and particularly strong robustness in collaboration-rich and long-sequence settings.

Overall, our results suggest that semantic alignment, together with hierarchical inductive structure, is a promising direction for advancing knowledge tracing in realistic educational environments. Future work includes extending MOSAIC to settings with explicit knowledge hierarchies and open-ended assessment, as well as developing more efficient strategies for integrating large language models into large-scale educational systems.